\documentclass[10pt,twocolumn,letterpaper]{article}

\usepackage[final]{cvpr}              % To produce the CAMERA-READY version

\usepackage[accsupp]{axessibility}  % Improves PDF readability for those with disabilities.

\usepackage{graphicx}
\usepackage{amsmath}
\usepackage{amssymb}
\usepackage{booktabs}

\usepackage[pagebackref,breaklinks,colorlinks]{hyperref}

\usepackage[capitalize]{cleveref}
\crefname{section}{Sec.}{Secs.}
\Crefname{section}{Section}{Sections}
\Crefname{table}{Table}{Tables}
\crefname{table}{Tab.}{Tabs.}

\usepackage{times}
\usepackage{epsfig}
\usepackage{graphicx}
\usepackage{amsmath}
\usepackage{amssymb}
\usepackage{epstopdf}
\usepackage{amsfonts,amssymb} 
\usepackage{amsmath, bm}
\usepackage{float}
\usepackage{tabu}
\usepackage{multirow}
\usepackage[table,xcdraw]{xcolor}
\usepackage{longtable}
\usepackage{booktabs}
\usepackage{setspace}
\usepackage{enumitem}
\usepackage{pifont}
\usepackage{tablefootnote}
\usepackage{color}
\usepackage{hyperref}
\usepackage{url}

\def\eg{\textit{e.g.}}
\def\etal{\textit{et al.}}

\def\ie{\textit{i.e.}}

\def\cvprPaperID{4919} % *** Enter the CVPR Paper ID here
\def\confName{CVPR}
\def\confYear{2022}

\begin{document}

%%%%%%%%% TITLE - PLEASE UPDATE
\title{SAR-Net: Shape Alignment and Recovery Network \\for Category-level 6D Object Pose and Size Estimation}

% \author{\hspace{-0.7cm}Jiashun Wang$^{1*\ddagger}$\quad Chao Wen$^{1*\mathsection}$\quad Yanwei Fu$^{1 \dagger\ddagger}$\quad Haitao Lin$^{1}$\quad Tianyun Zou$^{1}$\quad Xiangyang Xue$^{1}$\quad Yinda Zhang$^{2}$\footnotemark[2]\\
% \\\
% $^1$Fudan University \qquad $^2$Google LLC}

\author{Haitao Lin$^{1 \mathsection}$\quad Zichang Liu$^{1}$\quad Chilam Cheang$^{1}$\quad Yanwei Fu$^{1 \dagger\ddagger}$\quad Guodong Guo$^{2}$\quad Xiangyang Xue$^{1 \dagger}$\\
\\
$^1$Fudan University \qquad $^2$IDL, Baidu Research
}

\maketitle

{
  \renewcommand{\thefootnote}%
    {\fnsymbol{footnote}}
  \footnotetext[2]{indicates corresponding author.}
  
 \footnotetext[4]{Haitao Lin is with Academy for Engineering and Technology, and Engineering Research Center of AI and Robotics, Shanghai, China.}
 \footnotetext[3]{Yanwei Fu is with the School of Data Science, Fudan University, and Fudan ISTBI—ZJNU Algorithm Centre for Brain-inspired Intelligence, Zhejiang Normal University, Jinhua, China. This work was supported in part by NSFC under Grant (No. 62076067), Shanghai Municipal Science and Technology Major Project (2018SHZDZX01).}
 
}

% As a general rule, do not put math, special symbols or citations
% in the abstract or keywords.
%%%%%%%%% ABSTRACT
\begin{abstract}
  \vspace{-0.2cm}Given a single scene image, this paper proposes a method of Category-level 6D Object Pose and Size Estimation (COPSE) from the point cloud of the target object, without external real pose-annotated training data. Specifically, beyond the visual cues in RGB images, we rely on the shape information predominately from the depth (D) channel. The key idea is to explore the shape alignment of each instance against its corresponding category-level template shape, and the symmetric correspondence of each object category for estimating a coarse 3D object shape. Our framework deforms the point cloud of the category-level template shape to align the observed instance point cloud for implicitly representing its 3D rotation. Then we model the symmetric correspondence by predicting symmetric point cloud from the partially observed point cloud. The concatenation of the observed point cloud and symmetric one reconstructs a coarse object shape, thus facilitating object center (3D translation) and 3D size estimation. Extensive experiments on the category-level NOCS benchmark demonstrate that our lightweight model still competes with state-of-the-art approaches that require labeled real-world images. We also deploy our approach to a physical Baxter robot to perform grasping tasks on unseen but category-known instances, and the results further validate the efficacy of our proposed model. Code and pre-trained models are available on the project webpage~\footnote{Project webpage. \url{https://hetolin.github.io/SAR-Net}}.
\end{abstract}

%%%%%%%%% BODY TEXT
\vspace{-0.2cm}
\section{Introduction}
\label{section:introduction}
Estimating accurate 6D poses of objects plays a pivotal role in the tasks of augmented reality~\cite{marchand2015pose}, scene understanding~\cite{sui2017sum}, and robotic manipulation~\cite{deng2020self, tremblay2018deep, collet2009object, manuelli2019kpam, teng2016surface}. 
However, most 6D pose estimation works~\cite{deng2020self, deng2021poserbpf, 2019densefusion, wada2020morefusion, 2019pvnet, 2020single, 2019pix2pose, li2018deepim, he2021ffb6d} assume exact 3D CAD object models at \textit{instance-level}, which unfortunately greatly limits their practical applicability in real-world applications. 
To this end, this paper studies the task of Category-level 6D Object Pose and Size Estimation (COPSE). Thus the model is trained only by category-level supervision, reducing reliance on the exact CAD model for each instance.

\begin{figure}[t]
\begin{center}
\includegraphics[width=0.95\linewidth]{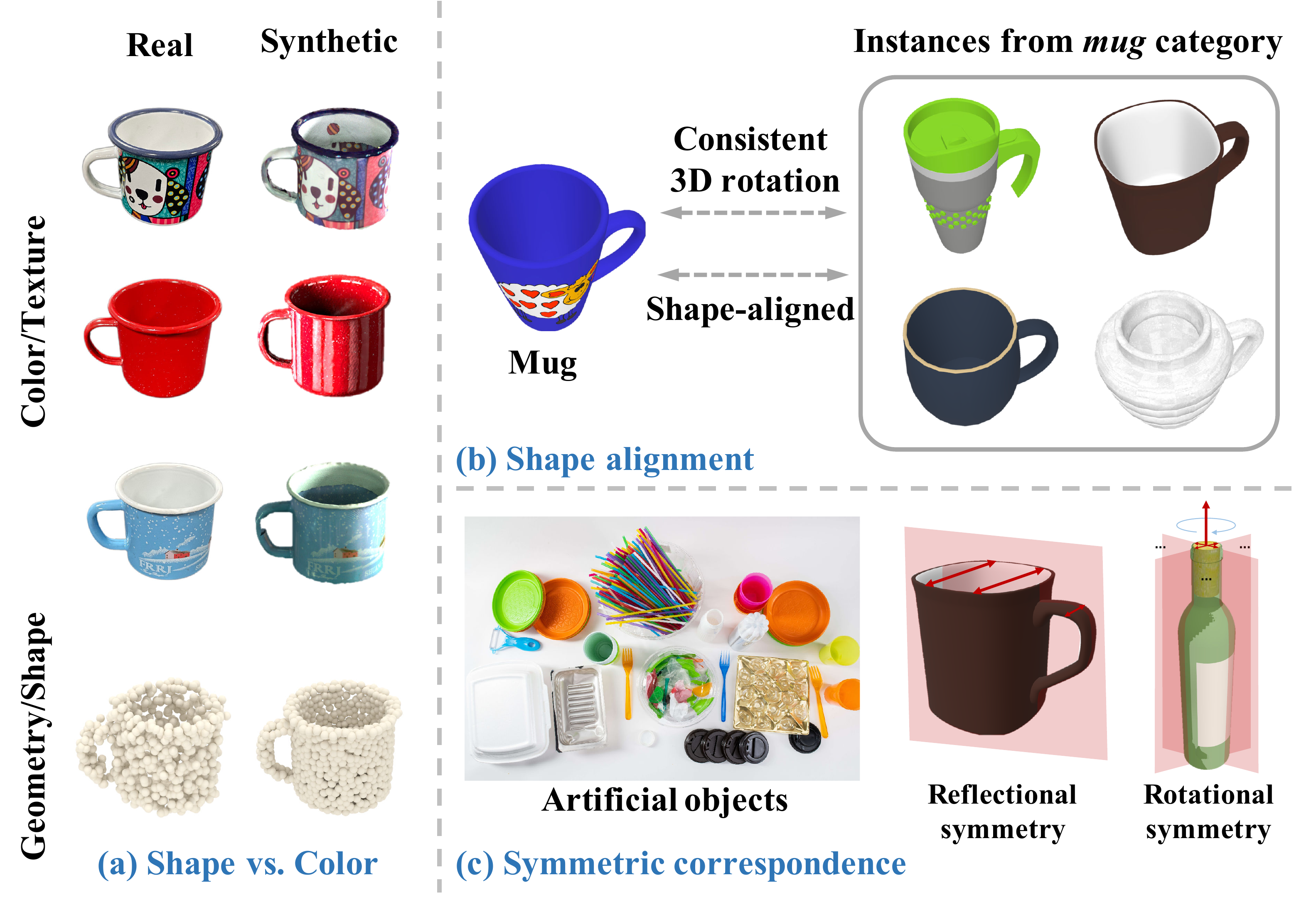}
\vspace{-0.25in}
\end{center}
   \caption{(a) The visual difference between synthetic and real images. They may have different textures and colors, but the shape and geometry maintain the same.
   (b) Illustration of shape alignment. Objects within a category have consistent 3D rotation if their shapes are visually aligned. (c) Illustration of symmetric correspondence.
   Most objects are manufactured with (near) symmetric shapes with reflectional symmetry or rotational symmetry. \label{fig:teaser}}
\vspace{-0.15in}
\end{figure}

Generally, the key challenge of  COPSE task lies in the huge color and shape variations of instances from the same  category~\cite{sahin2019instance, sahin2020review, sahin2018category}.
To handle \textit{intra-class variations}, previous works~\cite{wang2019normalized,chen2020learning, tian2020shape, lee2021category}
learn the RGB(-D) features of each instance to help map these instances into a unified space and minimize the intra-class variations. On the other hand, as the COPSE task relies on supervised learning from large amounts of well-labeled data, recent works~\cite{wang2019normalized,chen2020learning, tian2020shape, Lin_2021_ICCV} utilize synthetic data to train the COPSE model. Unfortunately, as illustrated in Fig.~\ref{fig:teaser}(a), the \textit{domain gap} between synthetic and real images potentially hinders the performance of COPSE model in the real-world deployment.

While most previous works exploit texture and color cues in RGB images, the shape information has been less touched, with some recent exceptions of reconstructing the observed point cloud~\cite{chen2021fs}, and
analyzing geometric stability of object surface patches~\cite{shi2021stablepose}.
For example, cups of similar or identical shapes have very diverse colors in Fig.~\ref{fig:teaser}(a). 
This motivates us to systematically explore shape information predominately from the depth (D) channel. Thus to alleviate challenges of intra-class variation and synthetic-real image domain gap, we propose encoding the shape by \textit{shape alignment} and \textit{symmetric correspondence}. Particularly,
our method encourages insightful shape analysis about geometrical similarity and symmetric correspondence.

\begin{figure}[t]
\begin{center}
\includegraphics[width=0.95\linewidth]{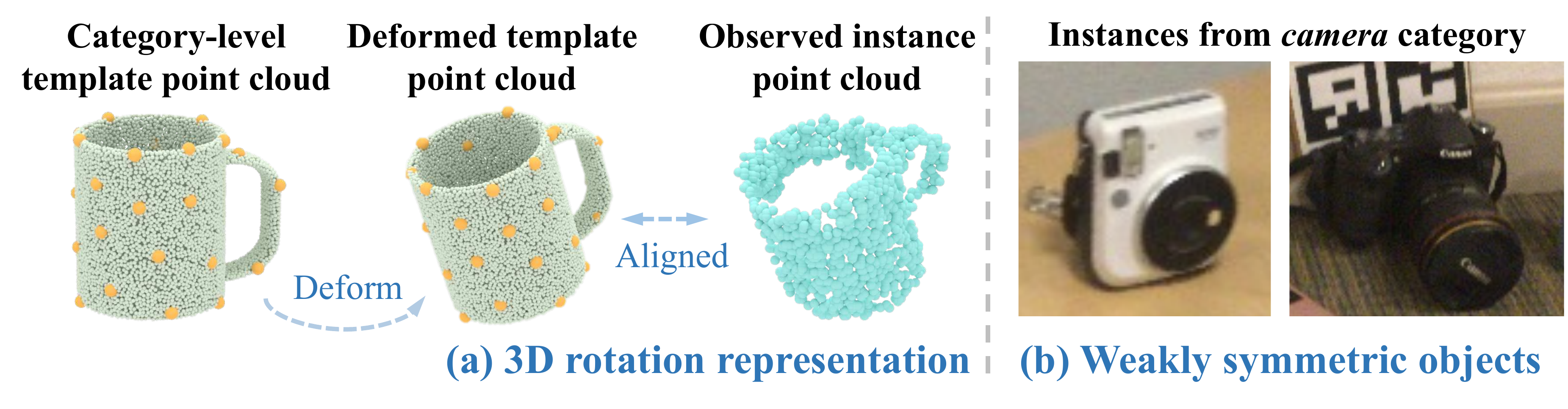}
\vspace{-0.25in}
\end{center}
   \caption{ 
   (a) Illustration of implicit 3D object rotation. The deformed template point cloud posses same 3D rotation with observed instance point cloud. (b) Illustration of weakly symmetric objects. Such objects usually have global symmetric shapes but asymmetric local parts.\label{fig:examplar_align}}
\vspace{-0.15in}
\end{figure}

\noindent \textbf{Shape alignment}.
Assuming the instances of the same category are well aligned by shapes, they should have the consistent 3D rotation, as cups are shown in Fig.~\ref{fig:teaser}(b). Thereby, the idea of shape alignment can be implemented as object 3D rotation consistency. In particular, given a category-level template shape in the form of a point cloud, it is deformed to align against the observed instance point cloud. We denote such the deformed template point cloud as an implicit representation for object 3D rotation, as shown in Fig.~\ref{fig:examplar_align}(a). 
Mathematically, the object rotation is thus recovered by solving the classical orthogonal Procrustes problem~\cite{schonemann1966generalized}, which calculates the approximation of alignment matrix between point clouds of the  category-level template and deformed one.
The shape alignment learns to be robust to intra-class variations of instances. 

\noindent \textbf{Symmetric correspondence}. 
Given the fact that many man-made object categories have the design principle with a symmetric structure~\cite{2020unsupervised}, symmetry is an important geometric cue to help our COPSE task. As in Fig.~\ref{fig:teaser}(c), the underlying symmetry allows for reasoning the reflectional and rotational symmetry of 3D shape from occluded 2D images. Note that specific object instances are practically never perfectly symmetric due to various shape variations of instances. To this end, we exploit the underlying symmetry by point clouds of objects, as our COPSE task does not demand the exact 3D shape recovery. Furthermore, we model the point cloud of symmetric objects by an encoder-decoder structure learned end-to-end with the other components of our framework. 
Thus, this actually facilitates the whole framework being robust to those objects which have some parts that are less symmetric as in Fig.~\ref{fig:examplar_align}(b). 

Formally, this paper proposes a novel Shape Alignment and Recovery Network (SAR-Net) to exploit the underlying object shapes for the COPSE tasks. Specifically, the RGB-D scene image is utilized as the input. We firstly employ Mask-RCNN~\cite{he2017mask} to pre-process the RGB image, and infer the segmentation mask and category label of each object instance. The points from depth channel are filtered by the predicted mask and further fed into the 3D segmentation network 3D-GCN~\cite{lin2020convolution} to generate  observed point cloud of the object. 
Furthermore, taking as inputs the point clouds of both observed object instance and category-level template, our SAR-Net predicts the implicit representation of deformed template point cloud, and infers symmetric point cloud. The 3D object rotation is further computed from the category-level and deformed template point clouds by Umeyama algorithm~\cite{umeyama}. 
Finally, we concatenate the observed and symmetric point clouds for a coarse object shape obtainment, which reduces the estimation uncertainty of object center (3D translation) and 3D size.
Extensive experiments conducted on the category-level NOCS dataset~\cite{wang2019normalized} demonstrate that our synthetic-only approach outperforms the state-of-the-art methods.

\noindent \textbf{Contributions}. Our main contribution is to propose a novel learning paradigm that efficiently encodes the shape information by the shape alignment and symmetric correspondence for the COPSE. We present a novel framework -- SAR-Net to implement this idea. In particular,  \\
\noindent 1) Based on shape similarities, our SAR-Net has the novel sub-net component that efficiently infers the implicit rotation representation
by shape alignment between point clouds of the category-level template shape and instance.\\
\noindent 2) A novel sub-net component for symmetric correspondence is proposed in this paper. It can predict symmetric point cloud from partially observed point cloud to obtain a coarse shape. The coarse shape helps to estimate the object center and size accurately.\\
\noindent 3) Practically, our SAR-Net is a very lightweight model with only 6.3M parameters. Such a single model is capable of doing the COPSE of multiple categories, and performs better than previous approaches of more model parameters. \\
\noindent 4) Critically, our SAR-Net is entirely trained on synthetic data and performs very well generalization on real-world scenarios. Remarkably, our synthetic-only approach still outperforms other competitors which typically require both synthetic and real-world data.

\begin{figure*}[t]
\begin{center}
\includegraphics[width=0.95\linewidth]{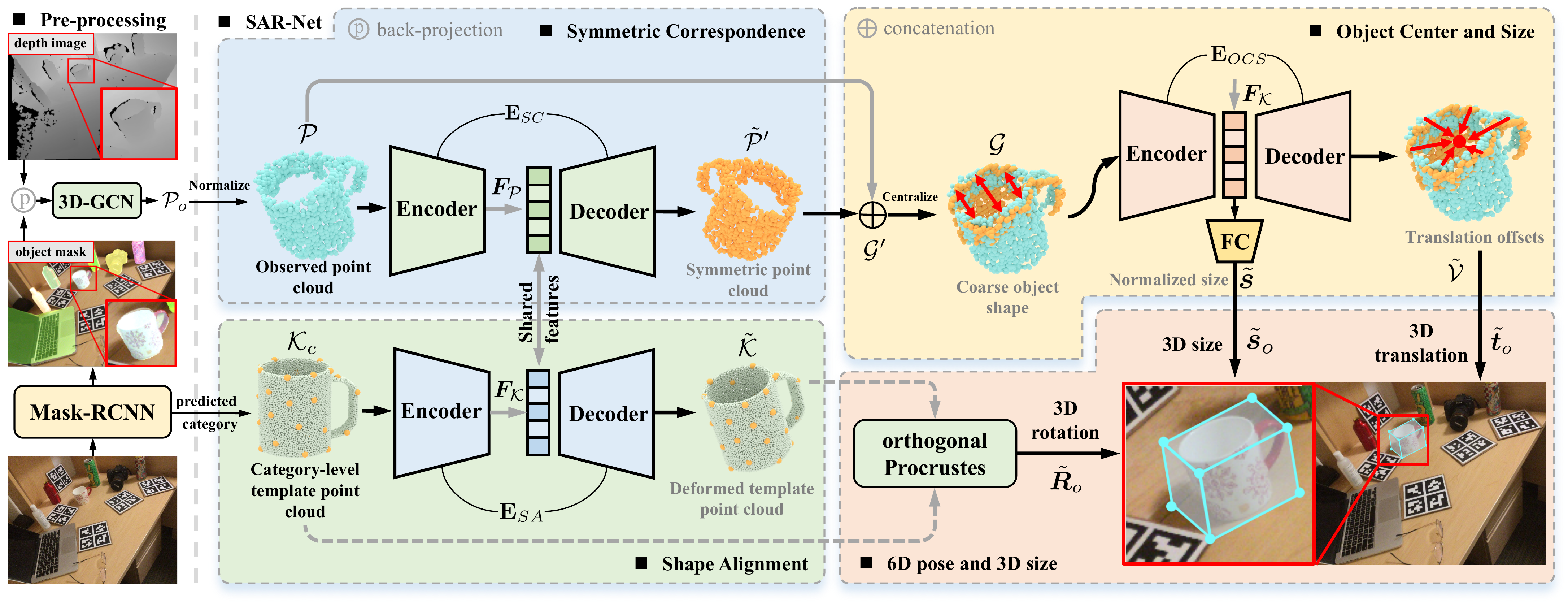}
\end{center}
\vspace{-0.3in}
   \caption{\small Architecture overview. The pre-processing stage (left) predicts the category and 2D segmentation mask of the target instance(\eg, \textit{mug}). 
   The points back-projected from depth channel are filtered by instance mask and processed by 3D-GCN to obtain object points $\mathcal{P}_o$ which are further normalized as $\mathcal{P}$.
  Our SAR-Net (right) takes both normalized point cloud $\mathcal{P}$ and  category-level template point cloud $\mathcal{K}_{c}$ as inputs to generate deformed template point cloud $\Tilde{\mathcal{K}}$ and symmetric point cloud $\Tilde{\mathcal{P}}'$, implemented by network $\mathbf{E}_{SA}$ and $\mathbf{E}_{SC}$, respectively.
   We get points of a coarse shape $\mathcal{G}$ by concatenating and centralizing the  $\mathcal{P}$ and  $\Tilde{\mathcal{P}}'$.
   From  $\mathcal{G}$, the network $\mathbf{E}_{OCS}$ predicts translation offsets $\Tilde{\mathcal{V}}$ and normalized size $\Tilde{\bm{s}}$. 
 The final 6D object pose $\{\Tilde{\bm{R}}_o, \Tilde{\bm{t}}_o\}$ and size $\Tilde{\bm{s}}_o$  are recovered by the post-processing stage in Sec.~\ref{section:3d_ocr} and Sec.~\ref{section:mdpv}. 
 \label{fig:architecture}} 
\vspace{-0.25in}
\end{figure*}

%%%%%%%%% Related Work
\section{Related Work}
\label{section:related_work}
\noindent \textbf{Instance-Level 6D Object Pose Estimation.} Most previous works~\cite{1999sift,2012linemod, 2011robustly, 2016textureless} estimate object pose by matching image features. Unfortunately, these methods are less efficient to infer poses of texture-less objects. 
Recent efforts are made on directly regressing 6D object pose from RGB images by CNN-based architectures, \eg, PoseNet~\cite{2015posenet} and PoseCNN~\cite{2017posecnn}.
DenseFusion~\cite{2019densefusion} introduces a cross-modal feature fusion manner for better aggregating color and depth information from RGB-D images, which infers more accurate objects pose than RGB-only methods. Such a fusion manner is also used in recent COPSE tasks~\cite{chen2020learning,tian2020shape}. 
Another line of works~\cite{2019cdpn,2018mheatmaps, 2017semantic, 2019pvnet, 2017bb8, 2018yolo6d, 2019dpod} first regress object coordinates or keypoints in 2D images and then recover poses by Perspective-n-Point algorithm~\cite{2009epnp}, \eg, PVNet~\cite{2019pvnet}. Recent approaches like~\cite{he2020pvn3d, he2021ffb6d} resort to 3D keypoint voting for precise pose estimation.
% Heuristically, PVN3D~\cite{he2020pvn3d} expends the keypoint voting mechanism from 2D to 3D space. 
In contrast to these keypoint voting methods~\cite{2019pvnet, he2020pvn3d, he2021ffb6d}, our approach focuses on a more practical setting without relying on exact object 3D models. 

\noindent \textbf{Category-Level 6D Object Pose Estimation.} Recent COPSE approaches~\cite{sahin2018category, wang2019normalized, chen2020learning, tian2020shape, shi2021stablepose,chen2021fs, lee2021category} vitally alleviate the limitation of previous instance-level tasks.
% Sahin~\etal~\cite{sahin2018category} first address the COPSE problem by training a part-based random forest with the 3D skeleton for pose recovery.
To handle the intra-class variations, most previous RGB-D methods ~\cite{wang2019normalized, chen2020learning, tian2020shape, lee2021category} represent instances of a category into a unified space.
Due to significant variations in object appearance, recent  methods~\cite{shi2021stablepose,chen2021fs, Lin_2021_ICCV} put more focus on geometric information of the object. StablePose~\cite{shi2021stablepose} is a depth-based method that analyzes geometric stability of object surface patches for 6D object pose inference.
% , and it also generalizes well for unseen instances. 
Lin~\etal~\cite{Lin_2021_ICCV} skillfully enforce the predicted pose consistency between an implicit pose encoder and an explicit one to supervise the training of the pose encoders and refine the pose prediction during testing. FS-Net~\cite{chen2021fs} extracts shape-based features from  point cloud of the target object for pose and size recovery. 
FS-Net estimates two perpendicular vectors for rotation decoupling. Compared to FS-Net, the representation of shape alignment transfers the rotation estimation problem into a reconstruction one. This representation has a more intuitive geometric meaning than FS-Net, as it provides visualization of aligned shape. Recent 6D pose trackers~\cite{wang20206, weng2021captra, wen2021bundletrack} achieve real-time tracking for category-level or novel objects with very good performance. Crucially, these methods have to rely on the good initial object pose and temporal information for the tracking. In contrast, the COPSE task addressed in this paper does not assume such a good initialization existed, and conducts the 6D object pose and size estimation from the single scene image.

\noindent \textbf{Symmetric Correspondence.}
Symmetric correspondence has been widely adopted in recent works~\cite{2017symmetry_wild, 2017reflection, 2015reflection}.
The reconstruction of symmetric objects has been investigated in~\cite{2011symmetry, 2014three}. Wu~\etal~\cite{2020unsupervised} use latent symmetric properties to disentangle components obtained from a single image. In the field of 6D pose estimation, HybridPose~\cite{2020hybridpose} is the first work to take the dense symmetric correspondences of an individual object as the intermediate representation to help the pose estimation. Differently, we fully utilize the symmetric correspondences in the same object category and extend 2D symmetric correspondences onto 3D ones, significantly improving COPSE inference performance.

%%%%%%%%% Method
\section{Methodology}
\label{section:method}

\noindent \textbf{Problem Formulation.} 
\label{section:problem formulation}
Given a depth image, segmented mask, and category of the target object, our goal is to estimate the 6D pose and 3D size of the object from its partially observed point cloud.
We represent the 6D object pose as a rigid-body homogeneous transformation matrix $\{\bm{R}_o,\bm{t}_o\} \in $ SE(3),
where 3D rotation $\bm{R}_o \in $ SO(3) and 3D translation $\bm{t}_o \in \mathbb{R}^3$. SE(3) and SO(3) indicate the Lie group of 3D rigid transformations and 3D rotation, individually. Finally, the 3D size of the object is formalized as $\bm{s}_o\in \mathbb{R}^3$.

\noindent \textbf{Overview.\label{section:overview}}
We give an overview of our SAR-Net, as in Fig.~\ref{fig:architecture}. Our method takes as input an RGB-D image. While RGB images are utilized by Mask-RCNN~\cite{he2017mask} in pre-processing stage to infer the segmentation mask and category of each instance, our SAR-Net only processes points from depth channel to address the COPSE task. Specifically, the points back-projected from depth channel are filtered by instance mask and processed by 3D segmentation network 3D-GCN~\cite{lin2020convolution} to obtain observed point cloud which is further normalized. (Sec.~\ref{section:preprpcessing}). 
The network $\mathbf{E}_{SA}$ is learned to deform the category-level template point cloud to align against the observed point cloud for 3D rotation representation (Sec.~\ref{section:3d_ocr}).
The symmetric correspondence is encouraged by the network $\mathbf{E}_{SC}$ to help predict the symmetric point cloud and complete the object shape (Sec.~\ref{section:GeoReS}). Finally, the object center and size are learned from the coarse shape by using the network $\mathbf{E}_{OCS}$ (Sec.~\ref{section:mdpv}).

\subsection{Pre-processing of  Point Cloud}
\label{section:preprpcessing}
\noindent \textbf{Processing observed point cloud.}
Given predicting segmented mask, we obtain the point cloud by back-projecting the masked depth. However, such a point cloud may still contain object and background points given by the imperfect segmentation. Thus, we further send this point cloud into the  3D-GCN~\cite{lin2020convolution} to purify the object points $\mathcal{P}_o \in \mathbb{R}^{3 \times N_{o}}$, where $N_{o}$ is the number of points in $\mathcal{P}_{o}$.
The 3D segmentation step makes our synthetically-trained model robust against the error from the 2D segmentation pipeline. Furthermore, we have to normalize the original observed point cloud $\mathcal{P}_{o}$. 
Particularly, we first calculate the centroid $\bm{r}_{o}=\sum{\mathcal{P}_{o} / N_{o}}$  of point cloud and maximum Euclidean distance $d_o = \max\{ \lVert \mathcal{P}_{o} - \bm{r}_{o} \rVert_2\ \}$ (scalar factor) relative to its centroid. We then normalize the $\mathcal{P}_{o}$ to obtain the  point cloud $\mathcal{P}$ by 
$ \mathcal{P} = (\mathcal{P}_o - \bm{r}_o)/d_o$.

\noindent \textbf{Processing category-level template point cloud.}
Given the 3D template dataset --  ShapeNetCore~\cite{chang2015shapenet},
we randomly\footnote{ SAR-Net is robust to random selection as in Appendix. } select one template per category as the \textit{category-level template shape}, which is   normalized by scale, translation, and rotation as in~\cite{wang2019normalized}.
Intuitively, instances of the same category should, at least in principle, have similar shapes as their category-level template shapes~\cite{kulkarni2019canonical}.
We further sample the category-level template shape into a sparse 3D point cloud $\mathcal{K}_{c} \in \mathbb{R}^{3 \times N_k}$ by using Farthest Point Sampling (FPS) algorithm~\cite{2019pvnet}, where $N_k$ is the number of points. 

\subsection{Shape Alignment}
\label{section:3d_ocr}
Given the normalized point cloud $\mathcal{P}$, our model learns the shape similarities among instances of the same category, deforming the category-level template point cloud $\mathcal{K}_{c}$ to align with the observed point cloud $\mathcal{P}$ visually, as demonstrated in Fig.~\ref{fig:examplar_align} (a). This module always reconstructs the 3D points in space of template point cloud but does not generate 3D points on the surface of observed point cloud, \ie, only transferring the rotational state of the observed point cloud $\mathcal{P}$ to deform the category-level template point cloud $\mathcal{K}_{c}$. 
Overall, it requires establishing a parametric encoder-decoder $\mathbf{E}_{SA}$ such that $\Tilde{\mathcal{K}} = \mathbf{E}_{SA}(\mathcal{K}_c, \mathbf{F}_{\mathcal{P}})$.
Then the task of rotation recovery $\Tilde{\bm{R}}_o$ is formulated  by the well-known orthogonal Procrustes problem~\cite{schonemann1966generalized} of alignment for two ordered sets of point clouds $\mathcal{K}_{c}$ and  $\Tilde{\mathcal{K}}$.

Concretely, our network uses a PointNet-like structure~\cite{qi2017pointnet, huang2020pf} as illustrated in Fig.~\ref{fig:architecture}. The normalized point cloud $\mathcal{P}$ and the category-level template point cloud $\mathcal{K}_c$ are fed into $\mathbf{E}_{SA}$ to extract shape-dependent features $\bm{F}_\mathcal{P}$ and $\bm{F}_\mathcal{K}$, respectively. 
We then concatenate $\bm{F}_\mathcal{P}$ and $\bm{F}_\mathcal{K}$ with each point in $\mathcal{K}_c$ to generate per-point feature embedding, thus performing shape-guided reconstruction of $\Tilde{\mathcal{K}}$ under the clues of geometric properties from observed point cloud $\mathcal{P}$. The reconstructed shape $\Tilde{\mathcal{K}}$ implicitly encodes the 3D rotation of the observed point cloud $\mathcal{P}$, as $\Tilde{\mathcal{K}}$ and $\mathcal{P}$ are enforced aligned by the network. We practivally obtain the ground-truth deformed template point cloud $\mathcal{K}$ by applying actual object rotation to the category-level template point cloud $\mathcal{K}_c$. Finally, the object 3D rotation is derived from Umeyama algorithm~\cite{umeyama} by solving registration of point clouds $\Tilde{\mathcal{K}}$ and $\mathcal{K}_{c}$ .

\begin{figure}[t]
\begin{center}
\includegraphics[width=0.95\linewidth]{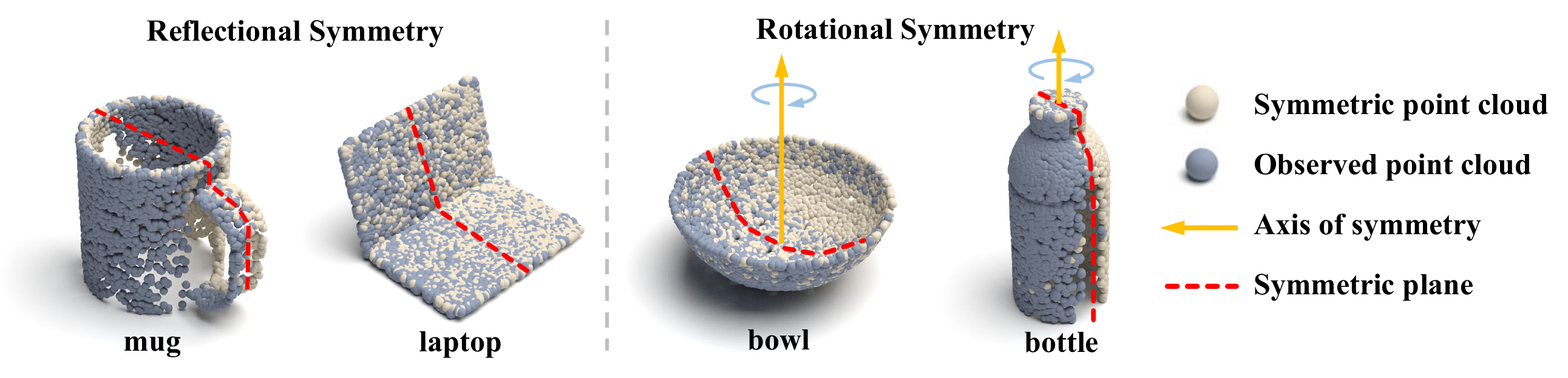}
\vspace{-0.25in}
\end{center}
   \caption{ 
   Illustration of symmetric point cloud. We generate the ground-truth symmetric point cloud from observed point cloud of  objects which have reflectional symmetry or rotational symmetry. \label{fig:examplar_symmetry}}
\vspace{-0.25in}
\end{figure}

\subsection{Symmetric Correspondence}
\label{section:GeoReS}
As most manufactured object categories have a symmetric structure,
we thus employ the reflectional and rotational symmetry as an essential geometric cue for the COPSE task. 
Such the underlying symmetry allows for reasoning the correspondence of potential symmetric point cloud from the observed point cloud.
We learn an encoder-decoder structure $\mathbf{E}_{SC}$ as the mapping function, to predict symmetric point cloud $\Tilde{\mathcal{P}}' = \mathbf{E}_{SC}(\mathcal{P}, \mathbf{F}_{\mathcal{K}})$ from observed point cloud $\mathcal{P}$. $\Tilde{\mathcal{P}}'\in \mathbb{R}^{3 \times N_o}$ and $\mathcal{P}$ have the same number of points. Concretely, We concatenate $\bm{F}_\mathcal{P}$ and $\bm{F}_\mathcal{K}$ with each point in $\mathcal{P}$ to generate point-wise feature embedding, and $\mathbf{E}_{SC}$ thus predicts corresponding symmetric point cloud $\Tilde{\mathcal{P}}'$ from $\mathcal{P}$. 

\noindent \textbf{Reflectional symmetry.}
As for object categories of reflectional symmetry like \textit{mug} and \textit{laptop}, they are usually symmetric around a fixed plane as shown in Fig.~\ref{fig:examplar_symmetry}. We treat this symmetry as a constraint of the prior symmetric plane to help complete object shape modeling. Thus, given observed points $\mathcal{P}$, we generate ground-truth symmetric points $\mathcal{P}'$ by flipping $\mathcal{P}$ along the symmetric plane. Thus, we present an encoder-decoder structure $\mathbf{E}_{SC}$ and learn to infer the corresponding symmetric points $\Tilde{\mathcal{P}}'$ to be symmetric with $\mathcal{P}$. 

\noindent \textbf{Rotational symmetry.} 
% \label{subsubsection:rotational}
Categories of rotational symmetry like \textit{bottle} and \textit{bowl} poss infinite symmetric planes around the axis of symmetry as in Fig.~\ref{fig:examplar_symmetry}, which hinders the network to get converged. 
One solution is to rotate observed points $\mathcal{P}$ by $180^\circ$ around its axis of symmetry in the object frame for generating ground-truth symmetric points $\mathcal{P}'$, in which case recovers the relatively complete object shape. Thus, the rotational symmetry is simplified as reflectional symmetry. It also enables our network $\mathbf{E}_{SC}$ to infer the occluded part from the observed point cloud to obtain a coarse shape for object center and size estimation. More examples of ground-truth symmetric points refer to Appendix.

\noindent \textbf{Remark.} 
It is noteworthy that specific object instances are never fully symmetric due to shape variations. Thus, exploiting the underlying symmetry by point clouds of objects is applicable to objects which have the global symmetric shapes but asymmetric local parts, as our framework does not demand the exact 3D shape recovery.

\subsection{Calculation of Object Center and Size \label{section:mdpv}}
Furthermore, we concatenate the predicted symmetric point cloud  $\Tilde{\mathcal{P}}'$ and observed point cloud $\mathcal{P}$ as in Fig~\ref{fig:architecture}. 
% The predicted symmetric point cloud and partial points are reflected before the concatenation operation.
This concatenation step generates a coarse 3D object shape $\mathcal{G}' \in \mathcal{R}^{3 \times 2N_o}$ for object center and size estimation. We then centralize points $\mathcal{G}'$ by using the calculated centroid $\bm{r}$ to get points $\mathcal{G}$, where $\bm{r}= \sum \mathcal{G}'/2N_{o}$. We use an encoder-decoder  $\mathbf{E}_{OCS}$ to infer the translation offsets  $\Tilde{\mathcal{V}}$ and normalized size $\Tilde{\bm{s}}$ from points in $\mathcal{G}$, \ie, $(\Tilde{\mathcal{V}},\Tilde{\bm{s}})=\mathbf{E}_{OCS}(\mathcal{G}, \mathbf{F}_\mathcal{K})$.
Notably, we incorporate ground-truth symmetric point cloud $\mathcal{P}'$ and the partial points $\mathcal{P}$ to prevent the unstable gradient propagation in the early training stage.

\noindent \textbf{Translation offset learning.} Inspired by previous 2D~\cite{2017posecnn, 2019pvnet} and 3D~\cite{he2020pvn3d, qi2019deep, he2021ffb6d} keypoint voting methods, we treat the object center as a specific keypoint. The encoder-decoder $\mathbf{E}_{OCS}$ infers 3D translation offsets $\Tilde{\mathcal{V}}=\{\Tilde{\bm{v}}_i\}_{i=1}^{2N_o}$, where $\Tilde{\bm{v}}_i$ denotes predicted translation offset from each point of $\mathcal{G}$ to the object center. The point cloud of coarse shape $\mathcal{G}$, together with predicted translation offsets $\Tilde{\mathcal{V}}$ votes for potential object center $\Tilde{\bm{t}}$.
Finally, the voted object center $\Tilde{\bm{t}}_o$ of the  observed point cloud $\mathcal{P}_{o}$ is given as below,
\vspace{-0.05in}
\begin{equation}
\begin{cases}
    \Tilde{\bm{t}} = \sum (\mathcal{G} + \Tilde{\mathcal{V}}) / 2N_o\\
    \Tilde{\bm{t}}_o = (\Tilde{\bm{t}} + \bm{r}) \cdot d_o + \bm{r}_o
\end{cases}
\label{equation:trans}
\vspace{-0.05in}
\end{equation}
\noindent where $d_o$ and $\bm{r}_o$ are scalar factor and centroid of observed point cloud $\mathcal{P}_o$ as computed in Sec.~\ref{section:preprpcessing}.

\noindent \textbf{Size estimation.} Obtained concatenated points $\mathcal{G}$, the network $\mathbf{E}_{OCS}$ regresses the normalized size $\Tilde{\bm{s}}$.
Then the actual size $\Tilde{\bm{s}_o}$ of the original point cloud $\mathcal{P}_o$ is recovered by the calculated scalar factor $d_o$ (Sec.~\ref{section:preprpcessing}), \ie, $\Tilde{\bm{s}}_o= d_o \cdot \Tilde{\bm{s}}$. Compared to regressing size from partially observed point cloud $\mathcal{P}$, the concatenated point cloud $\mathcal{G}$ provides a coarse shape for more accurate size estimation as discussed in Sec.~\ref{subsubsection:rough shape ablation}.

\subsection{Loss Function}
% \noindent \textbf{Associate multi-task loss}
\label{section:loss_func}
We  define the loss function $\mathcal{L}$ as follows,
\vspace{-0.08in}
\begin{equation}
\label{equation:mulitask}
    \mathcal{L} = \mathcal{L}_{def} +  \mathcal{L}_{sym} + \mathcal{L}_{cen} + \mathcal{L}_{size}
    \vspace{-0.08in}
\end{equation}
\noindent \textbf{Deformed point reconstruction loss.}
Our SAR-Net performs shape-guided reconstruction from observed points. Given ground-truth deformed template points $\mathcal{K}=\{\bm{k}_i\}_{i=1}^{N_k}$ with ${N_k}$ points, SAR-Net reconstructs $\Tilde{\mathcal{K}} = \{\Tilde{\bm{k}}_i\}_{i=1}^{N_k}$ as:
\vspace{-0.08in}
\begin{equation}
    \mathcal{L}_{def} = \frac{1}{N_k} \sum_{i=1}^{N_k} {\left\| \bm{k}_i -\Tilde{\bm{k}}_i \right\|_1}
    \vspace{-0.08in}
\end{equation}
For object with rotational symmetry, we adopt the strategy as in~\cite{wang2019normalized}. Refer to Appendix for details.

\noindent \textbf{Symmetric point reconstruction loss.}
The symmetric correspondence component predicts point-wise symmetric points $\Tilde{\mathcal{P}}'=\{\Tilde{\bm{p}}'_i\}_{i=1}^{N_o}$ based on the input observed points $\mathcal{P}$. We optimize the objective as:
\vspace{-0.08in}
\begin{equation}
    \mathcal{L}_{sym} = \frac{1}{N_o} \sum_{i=1}^{N_o}\left\| \bm{p}'_i - \Tilde{\bm{p}}'_i  \right\|_1
    \vspace{-0.08in}
\end{equation}
\noindent where $N_o$ is the number of points in $\mathcal{P}$; $\bm{p}'_i$ and $\Tilde{\bm{p}}'_i$ are the ground-truth and predicted symmetric points, respectively. 
As we simplify the rotational symmetry as a particular case of the reflectional symmetry (Sec.~\ref{section:GeoReS}). Thus, this loss function is also helpful to tackle the case of rotational symmetry.

\noindent \textbf{Translation offset loss.}
The network learns translation offsets $\{\Tilde{\bm{v}}_i\}_{i=1}^{2N_o}$ from concatenated points to object center. The learning of $\bm{v}_i$ is supervised by minimizing the loss as:   
\vspace{-0.05in}
\begin{equation}
    \mathcal{L}_{cen} = \frac{1}{2N_o} \sum_{i=1}^{2N_o} \lVert \bm{v}_i - \Tilde{\bm{v}}_i \rVert_1
    \vspace{-0.05in}
\end{equation}

\noindent where $2N_{o}$ is the number of concatenated points. The  $\bm{v}_i$ and $\Tilde{\bm{v}}_i$ are the ground-truth and predicted translation offsets.

\noindent \textbf{Size loss.}
For better size recovery, we regress the size from the point cloud of a coarse shape $\mathcal{G}$ as discussed in Sec.~\ref{section:mdpv}. 
We supervise the size regression as:
\vspace{-0.05in}
\begin{equation}
    \mathcal{L}_{size} = \left\| \Tilde{\bm{s}} - {\bm{s}} \right\|_{1}
    \vspace{-0.05in}
\end{equation}
\noindent where $\bm{s}$ and $\Tilde{\bm{s}}$ represent the ground-truth and predicted size of the normalized point cloud $\mathcal{P}$, respectively.

%%%%%%%%% Experiments
\section{Experiments}
\label{section:expriments}
\noindent \textbf{Datasets.}
(1) \textit{NOCS Dataset}~\cite{wang2019normalized}. It contains six object categories including \textit{bottle}, \textit{bowl}, \textit{camera}, \textit{can},  \textit{laptop}, and \textit{mug}. The NOCS has two parts, \ie, the synthetic part and the real-world one. For the synthetic part, there are 300K composite images, where 25K are set aside for evaluation(CAMERA25). For the real-world part, it contains 2.75K real-scene images for evaluation(REAL275). 
(2) \textit{LINEMOD Dataset}~\cite{2012linemod}. It is a widely used dataset for instance-level object pose estimation. It provides a scanned CAD model for each object.
\noindent(3) \textit{Additional Real-world Scenes.} Our model is tested on additional 6 different real scenes with 25 unseen instances from categories including \textit{bowl}, \textit{mug}, \textit{bottle}, and \textit{laptop}. The images are captured by a RealSense D435 camera and not manually pose-annotated.

\noindent \textbf{Evaluation Metrics.} 
\noindent \textit{(1) Category-level pose and size estimation.}
As~\cite{tian2020shape}, we compute average precision of 3D Intersection-Over-Union(IoU) at threshold values of 25\%, 50\% and 75\% for 3D object detection. 
The average precision at $\bm{m}^{\circ}\bm{n}cm$ is calculated for evaluating 6D pose recovery, \ie, the percentage of poses where the translational error is below $\bm{n}cm$ and the angular error is below $\bm{m}^{\circ}$. Here we choose threshold values of $5^{\circ}2cm$, $5^{\circ}5cm$, $10^{\circ}2cm$, and $10^{\circ}5cm$, respectively. 
\noindent \textit{(2) Instance-level pose estimation.}
We use average distance metric ADD~\cite{hinterstoisser2012model} for non-symmetric objects and ADD-S~\cite{2017posecnn} for symmetric objects (\eg, \textit{eggbox} and \textit{glue}). The accuracy of average distance less than 10\% of the object diameters is reported.

\noindent \textbf{Implementation Details.} 
 The architecture of SAR-Net and training details of 3D-GCN are presented in Appendix. 
We pick object models of six categories from ShapNetCore~\cite{chang2015shapenet} and utilize the Blender software~\cite{Blender} to render depth images to train our model, denoted as SAR-Net(small).
Additionally, we use 275K images from CAMERA dataset to train our model, denoted as SAR-Net. The training data rendered by Blender ($\sim$60K instances) is 10 times less than that of CAMERA dataset ($\sim$600K instances). The back-projected points of instances from synthetic depth images are disorderly sampled into 1024 points. 
Our SAR-Net is trained for 100 epochs with a batch size of 32 on a single RTX 2080Ti GPU. We initially set the learning rate as 0.0004 and multiply it by a factor of 0.75  every four epoch. We use the segmentation results of Mask-RCNN provided by~\cite{Lin_2021_ICCV} for fair comparisons. In robotic experiments, our SAR-Net is implemented on the desktop with an NVIDIA RTX2070 GPU, and pose and size estimation takes about 100ms. The model will be released on the repository of Baidu PaddlePaddle.

\subsection{Main Results}
\label{section:results}
\begin{figure*}[]
\begin{center}
\includegraphics[width=0.92\linewidth]{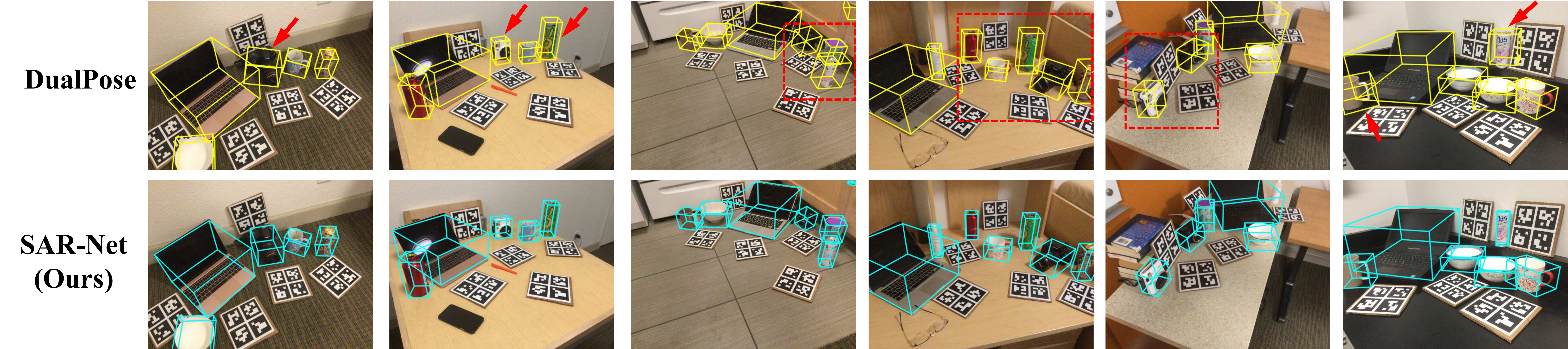}
\vspace{-0.25in}
\end{center}
   \caption{Qualitative comparisons between our SAR-Net and DualPose~\cite{Lin_2021_ICCV} on REAL275 dataset~\cite{wang2019normalized}. We visualize the estimated 6D pose and size as the tight-oriented bounding box around the target instances. \label{fig:qualitative}}
\vspace{-0.15in}
\end{figure*}

\noindent \textbf{Category-level NOCS Dataset.}
We compare our SAR-Net with NOCS~\cite{wang2019normalized}, CASS~\cite{chen2020learning}, SPD~\cite{tian2020shape}, FS-Net~\cite{chen2021fs}, StablePose~\cite{shi2021stablepose} and DualPose~\cite{Lin_2021_ICCV} on the CAMERA25 and REAL275 datasets in Tab.~\ref{tab:comparison_baseline}. 
For the synthetic CAMERA25 dataset, our SAR-Net achieves comparable performance to the state-of-the-art method DualPose and shows better performance under the more strict metric $5^{\circ}2cm$. 
For the real-world REAL275 dataset, NOCS, CASS, SPD and DualPose use both synthetic data (CAMERA) and real-world data for training; FS-Net and StablePose only use real-world data. In contrast, our method only uses synthetic data. Surprisingly, even in such a comparison, our SAR-Net outperforms all other baseline methods at all but $IoU$ metric, as FS-Net uses the pre-calculated mean size per category. The results validate the good generalization of SAR-Net in real-world applications. 
Although SAR-Net(small) is trained by using 10 times less synthetic training data than CAMERA, and it already outperforms all other methods but the DualPose on the REAL275 dataset. It could save memory footprint and reduce the training time in practice.
Moreover, for the pose and size estimation part of the model, our SAR-Net has less parameters than other methods. We also qualitatively show some results of our SAR-Net and DualPose~\cite{Lin_2021_ICCV} in Fig.~\ref{fig:qualitative}. Our method generates more accurate rotation estimation than DualPose, especially for the \textit{camera} category. More comparison results are shown in Appendix.
 
\begin{table*}[]
\centering
\renewcommand{\arraystretch}{0.9}
\footnotesize
\caption{\centering{Results on REAL275 and CAMERA25~\cite{wang2019normalized}: comparisons with other COPSE methods.($\uparrow$): higher better, ($\downarrow$): lower better.} \label{tab:comparison_baseline}}
\vspace{-0.1in}
\begin{tabular}{
c|c|cccccc|c|c}
\toprule[1pt]%\hline
 &  & \multicolumn{6}{c|}{mAP ($\uparrow$)} & Accuracy ($\uparrow$) &  Parameters ($\downarrow$) \\
\multirow{-2}{*}{Dataset} & \multirow{-2}{*}{Method} & $IoU_{50}$ & $IoU_{75}$ & $5^{\circ}2cm$ & $5^{\circ}5cm$ & $10^{\circ}2cm$ & $10^{\circ}5cm$ & $5^{\circ}5cm$ & (M) \\ \midrule[0.6pt]%\hline
 & NOCS~\cite{wang2019normalized} & 78.0 & 30.1 & 7.2 & 10.0 & 13.8 & 25.2 & 18.2 & - \\
& CASS~\cite{chen2020learning} & 77.7 & - & - & 23.5 & - & 58.0 & - & 47.2 \\
%  & 6-PACK~\cite{wang20206} & - & - & - & - & - & - & 33.3 & 81.5 $\times$ 6 \\
& SPD~\cite{tian2020shape} & 77.3 & 53.2 & 19.3 & 21.4 & 43.2 & 54.1 & 30.4 & 18.3 \\

& FS-Net~\cite{chen2021fs} & \textbf{92.2} & 63.5 & - & 28.2 & - & 60.8 & - & 41.2 \\

 & StablePose~\cite{shi2021stablepose} & - & - & - & - & - & - & 38.8 & - \\

& DualPose~\cite{Lin_2021_ICCV} & {79.8} & {62.2} & {29.3} & {35.9} & {50.0} & {66.8} & {50.1} & 67.9 \\ \cmidrule{2-10} 

& \cellcolor[HTML]{EFEFEF} SAR-Net(small) & \cellcolor[HTML]{EFEFEF} {80.4} & \cellcolor[HTML]{EFEFEF}\textbf{63.7} & \cellcolor[HTML]{EFEFEF}{24.1} & \cellcolor[HTML]{EFEFEF}{34.8} & \cellcolor[HTML]{EFEFEF}{45.3} & \cellcolor[HTML]{EFEFEF}{67.4} & \cellcolor[HTML]{EFEFEF}{49.1} & \cellcolor[HTML]{EFEFEF}\textbf{6.3} \\

\multirow{-8}{*}{REAL275} & \cellcolor[HTML]{EFEFEF}SAR-Net & \cellcolor[HTML]{EFEFEF}{79.3} & \cellcolor[HTML]{EFEFEF}{62.4} & \cellcolor[HTML]{EFEFEF}\textbf{31.6} & \cellcolor[HTML]{EFEFEF}\textbf{42.3} & \cellcolor[HTML]{EFEFEF}\textbf{50.3} & \cellcolor[HTML]{EFEFEF}\textbf{68.3} & \cellcolor[HTML]{EFEFEF}\textbf{54.9} & \cellcolor[HTML]{EFEFEF}\textbf{6.3}  \\
\midrule[0.6pt]

 & NOCS~\cite{wang2019normalized} & 83.9 & 69.5 & 32.3 & 40.9 & 48.2 & 64.6 & 49.4 & - \\
 & SPD~\cite{tian2020shape} & \textbf{93.2} &  {83.1} & 54.3 & 59.0 & 73.3 & {81.5} & 71.5 & 18.3 \\

& DualPose~\cite{Lin_2021_ICCV} & 92.4 & \textbf{86.4} & {64.7} & {70.7} & \textbf{77.2} & \textbf{84.7} & {79.9} & 67.9 \\ \cmidrule{2-10} 
& \cellcolor[HTML]{EFEFEF}SAR-Net(small) & \cellcolor[HTML]{EFEFEF}88.1 & \cellcolor[HTML]{EFEFEF}71.1 &\cellcolor[HTML]{EFEFEF}{44.0} & \cellcolor[HTML]{EFEFEF}{49.4} & \cellcolor[HTML]{EFEFEF}{56.1} & \cellcolor[HTML]{EFEFEF}65.6 & \cellcolor[HTML]{EFEFEF}{61.7} & \cellcolor[HTML]{EFEFEF}\textbf{6.3} \\ 

\multirow{-5}{*}{CAMERA25} & \cellcolor[HTML]{EFEFEF}SAR-Net & \cellcolor[HTML]{EFEFEF}86.8 & \cellcolor[HTML]{EFEFEF}79.0 & \cellcolor[HTML]{EFEFEF}\textbf{66.7} & \cellcolor[HTML]{EFEFEF}\textbf{70.9} & \cellcolor[HTML]{EFEFEF}{75.3} & \cellcolor[HTML]{EFEFEF}80.3 & \cellcolor[HTML]{EFEFEF}\textbf{81.4} & \cellcolor[HTML]{EFEFEF}\textbf{6.3}
\\\bottomrule[1pt]%\hline
\end{tabular}
\vspace{-0.15in}
\end{table*}

\noindent \textbf{Instance-level LINEMOD Dataset.} Our COPSE model could be easily employed in the instance-level task by using the exact object model as the template shape. We then generate 30K synthetic depth images for each instance for the model training. Compared with RGB(-D) methods~\cite{he2021ffb6d, 2019pvnet} or depth-only method~\cite{gao20206d, gao2021cloudaae}, our SAR-Net achieves comparable results in terms of ADD(-S) metric as in Tab.~\ref{tab:linemod}. Taking weakly symmetric objects (\eg,~\textit{cat}) as symmetric ones, our model still gains desirable performance, which provides evidence to support the utilization of symmetric correspondence to handle those objects which have some parts that are less symmetric.

\begin{table}[htb]
\centering
\footnotesize
\renewcommand\tabcolsep{3.0pt}
\renewcommand\arraystretch{0.9}
\caption{\centering{Results on LINEMOD~\cite{2012linemod}: comparisons with other instance-level methods. `S' is synthetic data and `R' is real data.}\label{tab:linemod}}
\vspace{-0.1in} 
\begin{tabular}{c|c|cccccc}
\toprule[1pt]
Training data & Methods & ape & can & cat & driller & eggbox & glue  \\ \midrule[0.6pt]
RGB(S+R) & PVNet~\cite{2019pvnet}   & 43.6  & 95.5 & 79.3 & 96.4 & 99.1 & 95.7 \\

% CAAE+ICP~\cite{2019densefusion} & 92.5 & 95.1 & 96.8 & 98.7 & 84.4 & 99.2 & 98.7 \\
RGBD(S+R) & FFB6D~\cite{he2021ffb6d} & \textbf{98.4} & \textbf{99.8} & \textbf{99.9} & \textbf{100.0} & \textbf{100.0} & \textbf{100.0}\\
D(S) &CP(ICP)~\cite{gao20206d} & 58.3 & 84.7 & 84.6 & 43.2 & 99.5 & 98.8\\ 
D(S) &CAAE~\cite{gao2021cloudaae} & {74.5} &{90.2} & 90.7 & {97.3} & {99.7} & 93.5\\ \cmidrule{2-8}  
D(S) & \cellcolor[HTML]{EFEFEF}SAR-Net & \cellcolor[HTML]{EFEFEF}64.5 & \cellcolor[HTML]{EFEFEF}83.6 & \cellcolor[HTML]{EFEFEF}{91.4} & \cellcolor[HTML]{EFEFEF}84.0 & \cellcolor[HTML]{EFEFEF}99.4 & \cellcolor[HTML]{EFEFEF}\textbf{100.0}\\
\bottomrule[1pt]
\end{tabular}
\vspace{-0.2in}
\end{table}

\noindent \textbf{Additional Real-world Scenarios.}
For \textit{additional real scenes} with multiple objects, visualization results are shown in Fig.~\ref{fig:realistic} (top row). Our model generates accurate estimation, as the objects are tightly located within predicted bounding boxes.
The results indicate the generalization capability of our SAR-Net in real-world applications, in terms of different depth sensors (\ie, Structure Sensor~\cite{StructureSensor} used by NOCS dataset, and RealSense D435 of ours) and different novel instances (18 novel instances from REAL275 and 25 novel ones of ours). See Appendix for more results.

\subsection{Ablation Studies}
We verify the efficacy of the key components of our SAR-Net on the REAL275 dataset in Tab.~\ref{tab:ablation}. 

\begin{table*}[]
\centering
\renewcommand\tabcolsep{3.0pt}
\renewcommand{\arraystretch}{0.9}
\footnotesize
\caption{\centering{Ablation studies of key components tested on REAL275. `SA' means shape alignment (Sec~\ref{section:3d_ocr}). `SC' means symmetric correspondence (Sec~\ref{section:GeoReS}). `Concat' indicates concatenating observed point cloud and symmetric one, and `Centralize' is the centralization operation (Sec.~\ref{section:mdpv}). `RegressT' means directly regressing object center versus predicting translation offsets in Sec.~\ref{section:mdpv}. `DSComp' is direct shape completion. 'Point Number' is the number of points in category-level template point cloud. (Sec.~\ref{section:preprpcessing}).
($\uparrow$): higher better.
}\label{tab:ablation}}
\vspace{-0.1in}
\begin{tabular}{c|cccccc|c|cccccc|c}
\toprule[1pt]
\rowcolor[HTML]{FFFFFF} 
 & \multicolumn{6}{c|}{Component} & Point & \multicolumn{6}{c|}{mAP ($\uparrow$)} & Accuracy ($\uparrow$) \\
\rowcolor[HTML]{FFFFFF} 
\multirow{-2}{*}{Row} & SA & SC & Concat & Centralize & RegressT & DSComp & Number & $IoU_{50}$ & $IoU_{75}$ & $5^{\circ}2cm$ & $5^{\circ}5cm$ & $10^{\circ}2cm$ & $10^{\circ}5cm$ & $5^{\circ}5cm$ \\ \midrule[0.6pt]%\hline
\rowcolor[HTML]{FFFFFF} 1 & \checkmark &  &  &   & & & 36 & 81.1 & 55.3 & 15.9 & 23.6 & 34.9 & 59.1 & 39.4 \\
\rowcolor[HTML]{EFEFEF} 
2  & \checkmark& \checkmark &  &  & &  & 36 & \textbf{81.2} & 60.1 & 17.8 & 27.7 & 38.8 & 63.3 & 44.1  \\
3 & \checkmark & \checkmark & \checkmark & &  & & 36 & 80.6 & 62.6 & 20.5 & 31.7 & 39.8 & 65.3 & 46.4 \\
\rowcolor[HTML]{EFEFEF} 
4 & \checkmark & \checkmark & \checkmark & \checkmark &  & & 36 & 80.4 & \textbf{63.7} & \textbf{24.1} & \textbf{34.8} & 45.3 & 67.4 & \textbf{49.1}  \\ \midrule[0.3pt]
\rowcolor[HTML]{FFFFFF} 5 & \checkmark & \checkmark & \checkmark & \checkmark & \checkmark & & 36 & 81.0 & 63.5 & 21.1 & 30.4 & 44.9 & 67.2 & 46.7  \\ 
\rowcolor[HTML]{EFEFEF} 
6 & \checkmark & &  & \checkmark &  & \checkmark & 36 & 80.6 & 59.5 & 19.2 & 28.9 & 41.6 & 65.2 & 45.0 \\ \midrule[0.3pt]
7 & \checkmark & \checkmark &  \checkmark & \checkmark & &  & 16 & 79.6 & 62.9 & 22.8 & 33.0 & \textbf{46.1} & \textbf{67.6} & 47.9 \\
\rowcolor[HTML]{EFEFEF} 
8 & \checkmark & \checkmark &  \checkmark & \checkmark & &  & 128 & 79.5 & 59.5 & 21.5 & 32.1 & 43.7 & 66.3 & 47.5  \\ \bottomrule[1pt]%\hline
\end{tabular}
\vspace{-0.15in}
\end{table*}

\noindent \textbf{Symmetric Correspondence.} 
We first check the importance of utilizing symmetric correspondence. We start from a basic network, which directly outputs the deformed template point cloud from shape alignment component (SA), translation offsets, and normalized size based on the \textit{partially observed point cloud}, as in {row 1}. The symmetric correspondence component (SC) is then added as shown in {row 2}.
The comparison results between \textbf{rows 1} and \textbf{2} illustrate that exploring underlying symmetric correspondence is a vital part of producing overall great results. 
With symmetry inference, the network learns to captures more useful shape characteristics for symmetric points reconstruction and also enhances the performance of other components.

\noindent \textbf{Partial or coarse shape.}
\label{subsubsection:rough shape ablation}
We then study the performance difference between using point cloud of a coarse shape and the partially observed point cloud.
We analyze it from two aspects:
(1) We concatenate the observed point cloud and symmetric one to obtain a coarse shape (Concat) for object center and size estimation, as in {row 3}. Comparing results in \textbf{rows 2} and \textbf{3}, estimation based on a coarse shape gains overall improved performance versus that relying on partially observed point cloud.
(2) We then explore the importance of centralization operation (Centralize) as in {row 4}, \ie, the concatenated point cloud is further centralized. The comparison results of \textbf{rows 3} and \textbf{4} indicate the necessity of the centralizing operation, yielding the further improvement of 1.1\% at $IoU_{75}$ and 3.6\% at $5^\circ2cm$, respectively.  

\noindent \textbf{Voting or regression for 3D center estimation.}
We replace the translation offset learning (Sec.~\ref{section:mdpv}) by regressing the object center (RegressT) as in row 5. The performance in \textbf{row 5} consistently drop under all metrics, comparing the results in \textbf{row 4}. Therefore, the translation offset learning and voting for object center help to locate a more accurate object center than RegressT.  

\noindent \textbf{Symmetry-based or direct shape completion.}
Also, we replace the SC component with direct object shape completion by using the same network (DSComp), supervised by the Chamfer loss~\cite{fan2017point}. Compared to results in \textbf{row 4}, performance of all metrics in \textbf{row 6} drop consistently. This is probably because the shape completion focuses on detailed reconstruction, which relies on a more complicated network, but inferring a symmetric point cloud is easier. Poor reconstruction results from a direct object completion network further degrade the performance of the object center and size estimation. 

\noindent \textbf{Point number of template point cloud.} 
In addition, We also explore the impact of the varied number of the category-level template point cloud $\mathcal{K}_c$ by using the full COPSE model. It is observed from \textbf{rows 4, 7, and 8} that 36 points are a good trade-off for our network to learn. The choice of 128 points degrades the performance due to the larger output space, while the 16 points are too sparse to represent the geometric structure of object, which negatively influences the final performance.  

We also conduct ablation studies of various rotation representation on the REAL275 dataset in Tab.~\ref{tab:3d-ocr}.\\ 
\noindent \textbf{Shape alignment for 3D rotation.}
We replace the shape alignment (SA) of SAR-Net by other 3D rotation representation in the form of quaternion, SVD~\cite{levinson2020analysis}, continuity 6D~\cite{zhou2019continuity} ($R_{6d}$), and Vector~\cite{chen2021fs}, respectively. The comparison results are summarized in Tab.~\ref{tab:3d-ocr}.
Compared to quaternion, SVD, and $R_{6d}$,  enforced shape alignment enables better generalization, as the point reconstruction search space is smaller than the rotation space, which is easier for the network to learn. The representation of Vector and SA both have geometric meaning, but our SA performs better than Vector, especially under strict metrics of $5^{\circ}2cm$ and  $5^{\circ}5cm$.

\begin{table}[t]
\centering
\footnotesize
\renewcommand\tabcolsep{1.0pt}
\renewcommand{\arraystretch}{0.9}
\caption{\centering{Ablation studies of using different 3D rotation representations tested on the REAL275 dataset. ($\uparrow$): higher better.}
\vspace{-0.15in}
\label{tab:3d-ocr}}
\vspace{-0.1in}
\begin{tabular}{c|cccccc|c}
\toprule[1.0pt]%\hline
 & \multicolumn{6}{c|}{mAP ($\uparrow$)} & Accuracy ($\uparrow$)\\
\multirow{-2}{*}{Method}   & $IoU_{50}$ & $IoU_{75}$ & $5^{\circ}2cm$ & $5^{\circ}5cm$ & $10^{\circ}2cm$ & $10^{\circ}5cm$ & $5^{\circ}5cm$         \\ \midrule[0.6pt] % \hline
quaternion & 80.6 & 62.9 & 20.8 & 29.7 & 43.6 & 64.6 & 46.3 \\
SVD~\cite{levinson2020analysis} & \textbf{82.2} & 61.8 & 17.8 & 24.3 & 39.6 & 58.6 & 40.0\\ 
$R_{6d}$~\cite{zhou2019continuity} & {81.6} & \textbf{64.1} & 21.7 & 30.5 & 42.6 & 64.2 & 46.6 \\
Vector~\cite{chen2021fs} & {81.2} & 62.5 & 21.1 & 31.5 & 45.1 & 67.1 & 47.6\\
\cmidrule{1-8}
\rowcolor[HTML]{EFEFEF} 
SAR-Net & 80.4 & 63.7 & \textbf{24.1} & \textbf{34.8} & \textbf{45.3} & \textbf{67.4} & \textbf{49.1} \\ \bottomrule[1.0pt]%\hline
\end{tabular}
\vspace{-0.15in}
\end{table}

\subsection{Robotic Experiments}
\label{section:robotic_grasp}
\begin{figure}
\begin{center}
\includegraphics[width=0.85\linewidth]{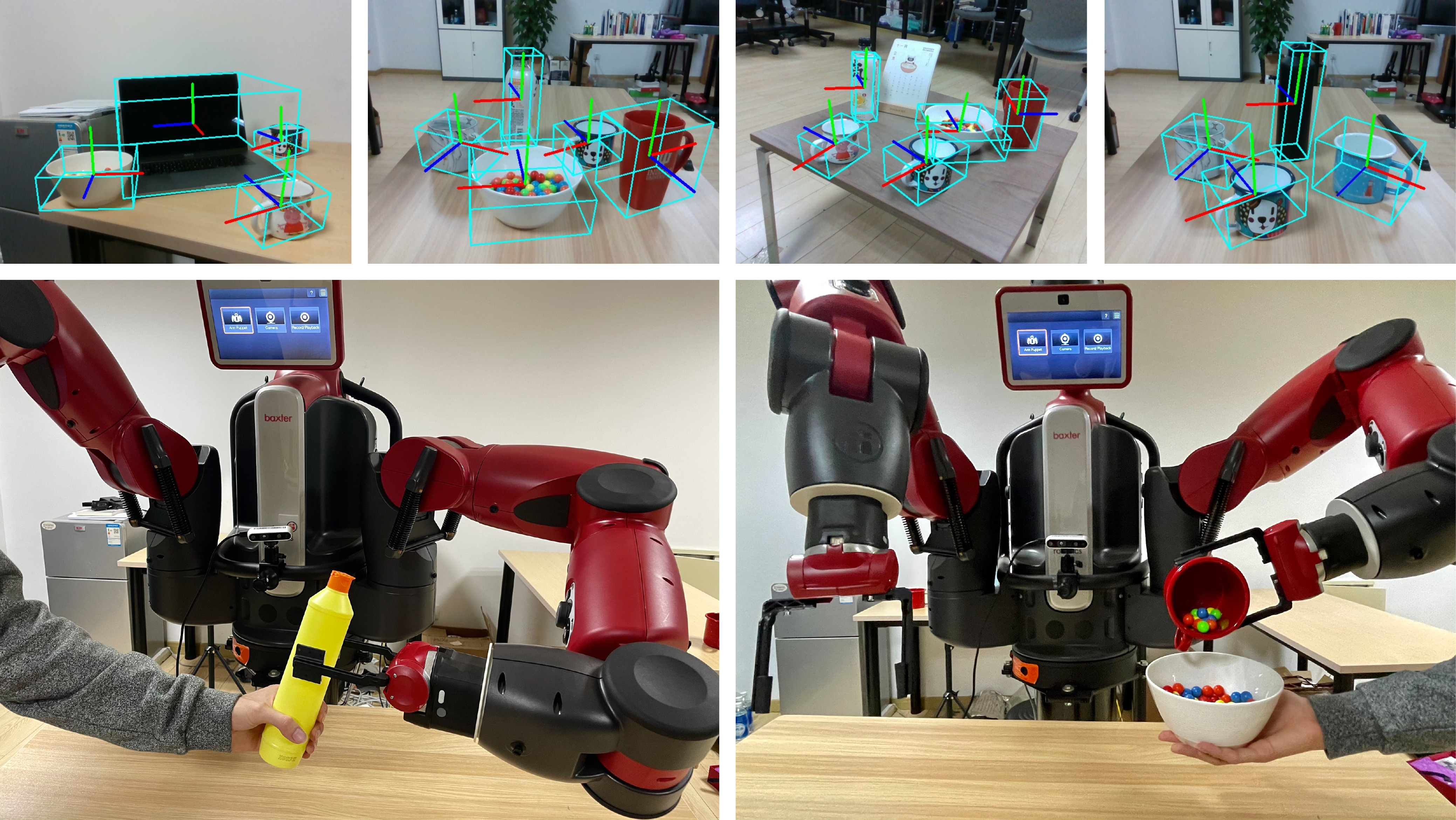}
\vspace{-0.25in}
\end{center}
  \caption{Estimated results given by our SAR-Net in various real clutter environments (top row). We perform tasks on a physical Baxter robot integrated with our SAR-Net (bottom row).\label{fig:realistic}}
\vspace{-0.15in}
\end{figure}

\noindent \textbf{Physic Baxter Robot.}
The robotic experiments compare the real-world performance of deploying COPSE models on a real Baxter robot executing different tasks, including object grasping, handover, and pouring as in Fig.~\ref{fig:realistic} (bottom row). Baxter is a dual-arm robot mounted with a RealSense D435 Camera on the base. More configurations of the robotic experiment are detailed in the Appendix.

\noindent \textbf{Grasping Task.} 
Particularly, we use 12 unseen instances from 3 classes, \ie, 4 mugs, 4 bottles, and 4 bowls. Deployed the COPSE models, the robot is programmed to attempt 10 grasps for each object.
In this experiment, our SAR-Net is compared against DualPose~\cite{Lin_2021_ICCV} and SPD~\cite{tian2020shape} with success rates of 88.3$\%$, 80.8$\%$ and 65.8$\%$, respectively. The baseline methods often fail due to imprecise rotation estimation or bigger estimated bounding boxes than the exact ones of target instances. See video demo for details.

\noindent \textbf{Object Handover Task.}
The robot interacts with the actor in this task, trying to grasp the objects in human hands. We choose the testing instance \textit{bottle}. 
The Baxter successes on 80$\%$ using our SAR-Net in 15 trials of the handover task, compared against that of 73.3$\%$ of DualPose and 66.7$\%$ of SPD, validating the accurate estimation of our SAR-Net.

\noindent \textbf{Pouring Task.} 
Using our COPSE model, we conduct the task of actor moving the bowl, while the robot follows the actor and executes pouring action. We choose each testing instance from \textit{bowl} and \textit{mug}, respectively. The robot is programmed to attempt 15 trials. Our SAR-Net is compared against DualPose and SPD with success rates of 73.3$\%$, 60.0$\%$, and 53.3$\%$, respectively.
The results show the efficacy of our COPSE model in robotic experiments. 

%%%%%%%%% Conclusion
\section{Conclusion}
\label{section:conclusion}
We propose a lightweight geometry-based model for the COPSE task. Our network uses shape alignment to facilitate 3D rotation calculation. The symmetry correspondence of objects is utilized to complete its shape for better object center and 3D size estimation. Our method achieves state-of-the-art performance without real-world training data. Furthermore, a physical Baxter robot integrated with our framework validates the utility in practical robotic applications. However, under the inherent limitation of the depth-based method, the sensor noise and lacked discriminative details may result in ambiguities in pose recovery. Future work will consider fusing additional color information from RGB channels for more accurate pose and size recovery.

% \section*{Acknowledgement}

\clearpage
%%%%%%%%% REFERENCES
{\small
\bibliographystyle{ieee_fullname}
\bibliography{egbib}
}

\end{document}